# ChatGPT v.s. Media Bias: A Comparative Study of GPT-3.5 and Fine-tuned Language Models

**Zehao Wen[1], Rabih Younes[2]**

[1]Shenzhen College of International Education, Shenzhen, 518043, China

[2]Duke University, Durham, NC 27704

[1]s21447.wen@stu.scie.com.cn, [2]rabih.younes@duke.edu

**Abstract.** In our rapidly evolving digital sphere, the ability to discern media bias becomes crucial as it can shape public sentiment and influence pivotal decisions. The advent of large language models (LLMs), such as ChatGPT, noted for their broad utility in various natural language processing (NLP) tasks, invites exploration of their efficacy in media bias detection. Can ChatGPT detect media bias? This study seeks to answer this question by leveraging the Media Bias Identification Benchmark (MBIB) to assess ChatGPT's competency in distinguishing six categories of media bias, juxtaposed against fine-tuned models such as Bidirectional and Auto-Regressive Transformers (BART), Convolutional Bidirectional Encoder Representations from Transformers (ConvBERT), and Generative Pre-trained Transformer 2 (GPT-2). The findings present a dichotomy: ChatGPT performs at par with fine-tuned models in detecting hate speech and text-level context bias, yet faces difficulties with subtler elements of other bias detections, namely, fake news, racial, gender, and cognitive biases.

## 1. Introduction

Media bias encompasses the selective presentation of content that favors a particular perspective, potentially influencing people's perceptions of events or issues [1]. The majority of Americans hold the belief that mass media organizations demonstrate bias [2]. Extensive research has focused on examining the influence of media bias, and it has been shown that this type of bias could exert a substantial impact on public opinion, that is, on elections and the societal reception of tobacco use. It also contributes to the dissemination of misleading information, impacting the decision-making process and eroding people's trust in news sources [1,3]. Therefore, detecting and understanding media bias has become crucial in the digital age, in which information is consumed rapidly and passively. Various methods have been proposed to identify bias in media, ranging from manual content analysis by human evaluators, e.g. [4, 5, 6, 7], to computational approaches employing machine learning and natural language processing techniques [8, 9, 10, 11]. However, these approaches typically focus on a particular type within the range of media bias, e.g., political bias or fake news, and suffer from challenges and limitations such as scalability and complexity of detecting linguistic nuances that contribute to bias.

In light of these limitations, the potential of Language Learning Models (LLMs) in understanding and generating text offers a new avenue for exploring the detection of media bias. LLMs, including transformer-based models like GPT, have been instrumental in revolutionizing several areas of natural language processing. ChatGPT from OpenAI is one such model[1]. Based on the architecture of Generative Pre-trained Transformers, ChatGPT utilizes GPT-3.5 as an engine through the process of adapting the model to human preferences. As per the official statement, ChatGPT possesses the ability to address subsequent inquiries, acknowledge any mistakes, challenge inaccurate assumptions, and diminish inappropriate demands due to its conversational character. Most significantly, ChatGPT has demonstrated considerable potential in various conventional natural language processing (NLP) tasks, including translation, sentiment analysis, reasoning, and summarization [12, 13, 14, 15].

While prior research has explored the application of human evaluation and AI models for media bias recognition, to the best of our knowledge, this is the first study to employ ChatGPT, a type of LLMs, for this purpose. We hope that our research will not only illuminate more about the capabilities of ChatGPT in identifying media bias, but it will also spark further investigation into the possibilities of AI in ensuring a balanced and accurate media landscape.

This paper proposes the use of ChatGPT to identify media bias across various spectra, testing its ability to understand and recognize biased language in a given text. The performance of ChatGPT is compared with fine-tuned language models like Bidirectional and Auto-Regressive Transformers (BART) [16], providing a comparative understanding of its strengths and weaknesses. ChatGPT is tested for identifying six types of bias, including racial bias, gender bias, cognitive bias, text-level context bias, hate speech, and fake news, giving an overview of ChatGPT's capability of dealing with different biases.

## 2. Related Work

This section reviews existing research on the detection of media bias, the evolution from manual to automated methods, and the capabilities and performance of large language models, specifically GPT-based models, in diverse tasks. These provide insights into the challenges and opportunities in detecting media bias and the potential role of GPT-based models in addressing these challenges.

*2.1. Identification of Media Bias*

Early approaches to detecting media bias relied primarily on human evaluation. For instance, Groseclose and Milyo evaluated media outlets' ideological leanings by comparing citation patterns with members of Congress' views, revealing a liberal bias in most of the outlets they examined [4]. A common technique of manual content analysis involves employing coders to systematically examine news texts and annotate sections of the texts they identify as manifesting media bias. To ensure consistency and precision in this process, codebooks containing definitions, detailed rules, and illustrative examples that provide clear instructions on how to annotate the texts are often utilized. For example, Papacharissi and Oliveira used a codebook to study framing orientations of coverage of terrorism in newspapers from the United States and the United Kingdom [5]. Similarly, Smith et al. developed a codebook to analyze the framing of protest events, suggesting that media often portray protests in ways that are counterproductive to social movement agendas. A codebook-based approach was applied to detect gender bias [6]. Van der Pas and Aaldering found that women received less attention and more negative coverage than men did, highlighting the importance of these methods in revealing different facets of media bias [10].

While human-based methods are instrumental in identifying media bias, their time- and labor-intensive nature often restricts the scope of research, making it difficult to analyze extensive datasets. As a result, automated approaches have emerged. For instance, D'Alonzo and Tegmark [8] developed a method to identify publishing newspapers based on automatic phrase frequency analysis, resulting in a probability distribution that allows for the automatic mapping of newspapers and phrases into a space representing bias. As computing power has surged, deep learning methods have become increasingly effective and popular for numerous NLP tasks. Thota et al. presented a dense neural network architecture for identifying fake news, using Tf-Idf vector representation, preprocessed engineered features, and cosine similarity measures as input features [11]. Their model demonstrated impressive performance on the test data. Attention mechanisms have also been utilized, as in the headline attention network proposed in [10], which achieved a high accuracy in detecting political bias. Lastly, Spinde et al. demonstrated the efficacy of a Transformer-based architecture trained on multiple bias-related datasets, achieving an impressive F1 score of 0.776, underlining the significant potential of automated bias detection methods [7].

---
[1] chat.openai.com/chat

However, current automated approaches face challenges and limitations that must be addressed for more comprehensive and accurate media bias detection. Primarily, existing models, while highly performant on specific tasks or datasets, may not generalize well across different types of media biases. Additionally, most automated methods rely heavily on labeled datasets for training, and the process of labeling data is not only resource-intensive, but also inherently subject to bias. This has led to growing interest in exploring unsupervised or semi-supervised methods for media bias detection. These challenges necessitate further research and improvement in the field of automated bias detection. This study is the first to evaluate whether pre-trained large language models can be used as an automated approach.

*2.2. Evaluation of the LLMs*

Language Language Models (LLMs) have emerged as powerful tools capable of generating and understanding human language by learning from extensive textual data. Among LLMs, GPT (Generative Pretrained Transformer) has emerged as a particularly influential model, showcasing impressive performance across diverse NLP tasks. For instance, Jiao et al. rigorously assessed the performance of ChatGPT in translation tasks against numerous benchmark test sets [12]. Their results highlighted that ChatGPT rivals other commercially available translation products, especially for high-resource European languages. Furthermore, investigations into ChatGPT's zero-shot reasoning capacity have been conducted, demonstrating competitive performance with fine-tuned models across many benchmarks, despite inferior results on some specific datasets [13]. In addition, ChatGPT has been shown to generate new summaries on par with humans and even surpass traditional fine-tuned models in terms of aspect-based summarization [13, 14]. Finally, ChatGPT, in its role as a zero-shot sentiment analyzer, matches the performance of fine-tuned BERT and other state-of-the-art models trained on domain-specific labeled data [15]. Given the demonstrated proficiency of ChatGPT in processing and understanding human language, it is compelling to explore its efficacy in detecting media bias within text. This examination reveals the capability of ChatGPT as a sufficient tool for fostering a more balanced media landscape, thus contributing to the ongoing academic discourse on enhancing objectivity in media reporting and consumption.

## 3. Experiment Setup

This section delineates our experimental setup, which includes the preparation and division of the Media Bias Identification Benchmark (MBIB) data for testing, the fine-tuning of three existing models for comparison against ChatGPT, and the design of prompts for eliciting ChatGPT's bias detection abilities across six distinct bias identification tasks. These elements are vital for understanding the experiment's procedure, validating the subsequent results, and enabling replication of our study.

*3.1. Test Data*

The analysis presented in this paper utilizes a subset of the Media Bias Identification Benchmark (MBIB) for test data. MBIB is a comprehensive dataset compiled by Wessel et al. that serves as a common framework for evaluating different media bias detection techniques [1]. Following a meticulous review of 115 datasets, the authors curated a selection of nine tasks and 22 associated datasets, specifically tailored to evaluate media bias detection techniques. The datasets within the MBIB undergo a specific preprocessing step to convert the labels into a binary format, simplifying the integration of different datasets by removing the need for diverse model heads and keeping the task formulation straightforward. For datasets with continuous labels, binarization is achieved by defining a threshold, with the author's recommended threshold followed wherever possible.

Owing to data availability, our study focuses on six of the nine tasks represented in the MBIB. Nonetheless, the selected tasks offer a broad assessment of ChatGPT's capabilities in detecting media bias. Table 1 introduces the six tasks involved in this work.

**Table 1.** the description of each bias that ChatGPT is tested on.

| Task | Description |
| --- | --- |
| Text-Level Context Bias | This task looks at how the context of a text can shape the reader's perspective through the use of words and statements. |
| Cognitive Bias | This task deals with the biases that arise from readers' selections of articles to read and the sources they choose to trust, which can be magnified through the influence of social media. |
| Hate Speech | This task includes the use of language that expresses animosity towards a particular group or seeks to patronize, embarrass, or provoke offense. |
| Fake News | This task pertains to the dissemination of published material that relies on false assertions and premises, presented as factual to mislead the reader. |
| Racial Bias | This task involves negative or positive portrayals of racial groups. |
| Gender Bias | This task encompasses the issue of gender-based discrimination in media, involving the underrepresentation or negative depiction of one gender. |

Given the varying sizes of the datasets within MBIB, each dataset is proportionally split into training and testing subsets for the purposes of this study. For most bias identification tasks, an 80-20 training-testing split is employed on the datasets. However, due to the large number of examples included in the cognitive bias and hate speech tasks, i.e., 2 million examples, we randomly selected 10% of each dataset pertaining to these tasks and perform the 80-20 training-testing split to this subset of data. Table 2 reveals the adopted size of each task.

**Table 2.** the adopted size for each bias identification task.

| Task | Training Size | Test Size |
| --- | --- | --- |
| Text-Level Context Bias | 7213 | 1805 |
| Cognitive Bias | 39066 | 9768 |
| Hate Speech | 27879 | 6972 |
| Fake News | 9063 | 2675 |
| Racial Bias | 7830 | 1958 |
| Gender Bias | 32072 | 8020 |

*3.2. Fine-Tuned Models*

In this study, we establish a point of comparison for evaluating the performance of ChatGPT in detecting media bias. Three well-regarded models are selected for this purpose: ConvBERT, BART, and GPT-2. These pre-trained language models, recognized for their strong performance across various Natural Language Processing (NLP) tasks, are briefly described below.

BART is a denoising autoencoder that maps both the original sequence and a corrupted version to the same latent space, allowing it to perform a variety of generation and understanding tasks [16]. Convolutional Bidirectional Encoder Representations from Transformers (ConvBERT) is a variation of BERT that incorporates convolutional layers into the architecture to capture local dependencies within the text [17]. Generative Pre-trained Transformer 2 (GPT-2), developed by OpenAI, is a large-scale transformer-based language model that generates coherent and contextually relevant sentences by anticipating the subsequent word in a sequence using the context provided by the preceding words [18].

We fine-tune them on each identification task with the split training set as mentioned in section 3.1, at a learning rate of $5e^{-5}$, batch size of 128, and epoch number of 5. The performance of these models on the test datasets serves as a comparison against the performance of ChatGPT.

*3.3. Prompt Engineering*

This research employs a specialized version of ChatGPT, the GPT-3.5-turbo, recognized for its proficiency and cost-effectiveness. To facilitate replicability and consistency in the outcomes, the temperature of the model has been set to zero, ensuring that the model functions in a deterministic fashion and produces identical responses to identical prompts. The methodology employed for generating queries to interact with the model is adapted from the methods used in [18]. We ask ChatGPT to provide three concise prompts for eliciting its full ability in identifying each of the six types of bias. For example, our query for identifying racial bias is:

Please give me three concise prompts for eliciting your full capability in identifying whether there is racial bias in a text given to you (i.e., whether the text includes negative or positive portrayals of racial groups).

Then, as suggested by [18], the three model-generated prompts for each identification task are tested on a limited set (i.e., 60 examples) derived from the original datasets. Given that a task incorporates multiple datasets, we have instituted a procedure for random selection of equivalent quantities of examples from each dataset. The chosen examples also include an identical number of positively labeled instances (signifying the presence of a specific type of bias) and negatively labeled instances (indicating the absence of such bias), all of which are excluded from the test set. Based on these outcomes, the most plausible and effective prompt is selected, with final selections presented in Table 3.

**Table 3.** The final prompts we used to test ChatGPT under each bias identification task.

| Bias Type | Prompt |
| --- | --- |
| Racial Bias | Identify any language or descriptions that may perpetuate stereotypes or reinforce negative perceptions of a particular racial group. |
| Gender Bias | Consider the overall message or theme of the text and evaluate whether it perpetuates any harmful stereotypes or reinforces gender roles. |
| Fake News | Does the text use emotionally charged language or appeal to personal biases? Are there any logical fallacies present in the argument presented? |
| Hate Speech | Consider the overall tone and intent of the text, and whether it appears to be promoting or inciting hatred or discrimination towards a specific group. |
| Cognitive Bias | What is the author's tone and perspective? Are they objective or subjective? Are there any underlying biases or prejudices evident in the language used? |
| Text-Level Context Bias | Identify any loaded language or emotionally charged words in the text and analyze how they may influence the reader's perspective. |

Additionally, a directive is attached behind the task prompt to ensure that the model responds in a manner that can be processed automatically. This directive requests the model to render its output in JSON format, incorporating a 'bias' column that denotes either 1 or 0, signifying the presence or absence, respectively, of bias within the provided text.

**4. Evaluation**

This section delves into a comprehensive examination of ChatGPT's performance in comparison to other fine-tuned models on six media bias identification tasks. This assessment is essential to understand the effectiveness of these models in recognizing and mitigating bias in various contexts, contributing to the development of more balanced AI systems. By analyzing the model performance using specific metrics, we offer insights into their strengths, limitations, and potential areas for improvement.

*4.1. Evaluation Metrics*

As recommended by the original authors of MBIB [1], two metrics are employed to assess the performance of ChatGPT compared with other fine-tuned models:

- *Micro Average F1-Score*: A single F1-score is deduced for the predictions made by a model on the complete test set. This method disregards the variation in the dataset from which each example is from. This metric gives an easy look into the overall performance of the model.

- *Macro Average F1-Score*: A F1-score is calculated solely for each individual dataset in the test set, and the results are averaged to obtain a macro average score. This approach ensures that all datasets contribute evenly to the final score regardless of their size.

The performances of ChatGPT and other fine-tuned models are presented in Table 4.

**Table 4.** The performance of models over each bias identification task, measured by two metrics: micro average f1-score and macro average f1-score.

| Bias Type | Model | Micro-Score | Macro-Score |
|---|---|---|---|
| Racial Bias | ChatGPT (zero-shot) | 0.6288 | 0.7207 |
|  | BART | 0.7873 | 0.8679 |
|  | ConvBERT | 0.7540 | 0.8268 |
|  | GPT-2 | 0.7792 | 0.8736 |
| Fake News | ChatGPT (zero-shot) | 0.5021 | 0.4346 |
|  | BART | 0.7060 | 0.6800 |
|  | ConvBERT | 0.6759 | 0.6844 |
|  | GPT-2 | 0.6739 | 0.6613 |
| Text-level Context Bias | ChatGPT (zero-shot) | 0.7445 | 0.4414 |
|  | BART | 0.7602 | 0.4083 |
|  | ConvBERT | 0.7873 | 0.3965 |
|  | GPT-2 | 0.7818 | 0.3941 |
| Hate Speech | ChatGPT (zero-shot) | 0.6929 | 0.8236 |
|  | BART | 0.8725 | 0.8697 |
|  | ConvBERT | 0.8784 | 0.8745 |
|  | GPT-2 | 0.8702 | 0.8390 |
| Gender Bias | ChatGPT (zero-shot) | 0.4945 | 0.5945 |
|  | BART | 0.8262 | 0.7824 |
|  | ConvBERT | 0.8263 | 0.7779 |
|  | GPT-2 | 0.8212 | 0.7776 |
| Cognitive Bias | ChatGPT (zero-shot) | 0.2362 | 0.2533 |
|  | BART | 0.6582 | 0.4711 |
|  | ConvBERT | 0.6673 | 0.5153 |
|  | GPT-2 | 0.6729 | 0.4846 |

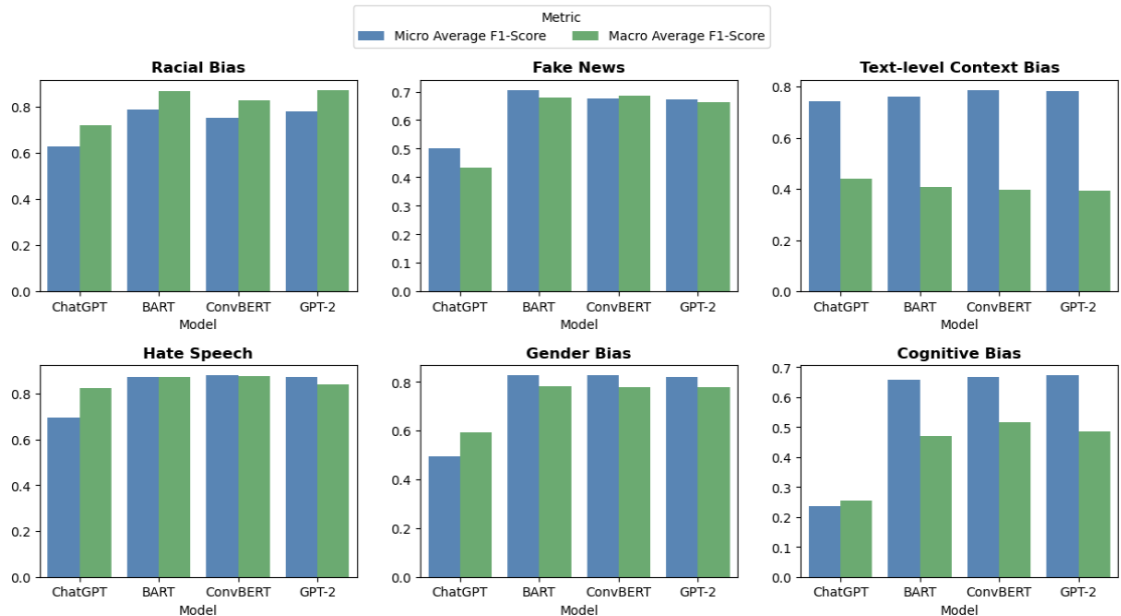

**Figure 1.** Bar charts showcasing the results of zero-shot ChatGPT and fined-tuned language models on the six media bias identification tasks.

The fine-tuned methods have similar performance, whereas the performance of ChatGPT varies significantly across tasks. The full data is attached at the appendix (Figure 1).

*4.2. Analysis*

Overall, zero-shot ChatGPT performs weakly on most bias identification tasks when compared to fine-tuned methods such as BART, ConvBERT, and GPT-2. In general, this disparity might be reflective of the inherent subjectivity involved in bias labeling. Despite the use of detailed codebooks to guide the labeling process, the interpretation of bias varies from one individual to another; what might be perceived as bias by one person may not be considered by another. Fine-tuned methods like BART, ConvBERT, and GPT-2 have the advantage of being explicitly trained to adapt to the patterns and subtleties of how human labelers identify bias, and therefore they achieve higher scores. In contrast, the zero-shot nature of ChatGPT limits its proficiency in bias identification, as it only depends on overall patterns within the extensive training data it is exposed to, without learning the way the constructors of the datasets label.

In terms of gender and racial bias, ChatGPT significantly underperforms the fine-tuned methods. Interestingly, ChatGPT tends to overestimate in these domains, marking a high number of false positives. This discrepancy arises when ChatGPT interprets content as biased, which human evaluators and other models classify as neutral. An example includes the statement, *"I can't stand a Yankee voice commentating on football. CRINGE,"* which ChatGPT labels as biased, explained by "It reinforces gender roles by assuming that football commentary is a male-dominated field". This over-sensitivity could be attributed to the exposure to bias during training, causing the model to associate specific stereotypes or biases with certain words or phrases, which, in this case, it associates the term "Yankee voice" with the assumption that football commentary is predominantly done by males.

In the realm of cognitive bias and fake news detection, ChatGPT struggles substantially, largely underperforming compared with fine-tuned methods such as BART, ConvBERT, and GPT-2. Even humans' scuffle to identify these biases. Cognitive biases are often contextually embedded or subtly expressed, without explicit words, compared to tasks such as hate speech, requiring a nuanced understanding of the subject matter and current events, which is difficult to encapsulate within a single-sentence input in a zero-shot learning context. Additionally, fake news adds to this challenge because of its ambiguous and often deceptive nature, making it difficult to distinguish from the truth based on language cues alone. Therefore, it is reasonable for ChatGPT to perform at this standard in a zero-shot manner.

However, ChatGPT demonstrates relatively competitive performance in detecting hate speech, although slightly trailing behind the fine-tuned models. Hate speech typically manifests through explicit, often aggressive language patterns and stark negative sentiments, making it more identifiable compared to subtler forms of bias. This comparative ease of detection illuminates ChatGPT's inherent capabilities in discerning explicit negativity and aggression despite the absence of task-specific fine-tuning. With regard to text-level context bias detection, ChatGPT exhibits results comparable to those of the fine-tuned methods. The good performance of ChatGPT in text-level context bias detection might stem from the benefits of its large-scale architecture. The extensive architecture is better equipped to understand language intricacies and subtle meanings inherent in human communication. The rich, multifaceted linguistic understanding that ChatGPT has gained through its comprehensive training enables it to decipher how contextual elements can influence conveyed meaning even in a zero-shot setting.

The noticeably low macro-average scores across all models for this bias type can be attributed to inconsistent model performance across the two available datasets under this task. Specifically, all models struggle to perform well on one dataset, which contains a small number of test examples (i.e., 100 examples), while showing better performance on the other. However, given the limited dataset quantity, the macro-average score, in this case, may not be an accurate representation of model performance.

In conclusion, while ChatGPT exhibits a certain proficiency, it may not serve as a definitive detector for media bias in its present form. However, it has been shown that few-shot prompting could potentially improve ChatGPT's performance across natural language processing tasks [18], but this is not done in this work considering the subjectivity and variability of bias detection. The hint given to ChatGPT through few-shot prompting may be inconsistent with other examples in the datasets. In addition, human evaluation may be carried out to test the effectiveness of ChatGPT in accurately identifying bias in real-life scenarios. This could serve to underscore the model's strengths and weaknesses, paving the way for understanding and fully eliciting its ability to help with a more balanced and healthy information system.

## 5. Conclusion

This research contrasts ChatGPT's proficiency in detecting various media biases with fine-tuned models BART, ConvBERT, and GPT-2. Although ChatGPT demonstrates notable success in identifying hate speech and text-level context bias, it underperforms in tasks requiring deeper contextual understanding, such as gender, racial, and cognitive biases. These results suggest that while large language models have made significant strides in language understanding, they still fall short in tasks that require a deeper, more nuanced understanding of context and bias. Other factors, such as the subjectivity of bias criteria and the data on which ChatGPT is trained, might also contribute to the disparities in zero-shot ChatGPT's performance with other fine-tuned models. The current study provides a stepping stone towards a deeper understanding of the role that large language models can play in ensuring a balanced and healthy information ecosystem. Future studies should investigate the enhancement of such models using techniques such as few-shot prompting and human evaluation, as strides towards a balanced information ecosystem.